\documentclass[10pt]{article} % For LaTeX2e

\usepackage[preprint]{rlj} % Should be uncommented for preprint versions

\usepackage{amssymb}            % Defines common symbols like \mathbb R
\usepackage{mathtools}          % Extends amsmath, providing common math tools
\usepackage{mathrsfs}           % Enables \mathscr, which can work in cases that \mathcal does not
\usepackage{graphicx}           % For including images
\usepackage{subcaption}         % Allows for the use of subfigures and subcaptions
\usepackage[space]{grffile}     % For spaces in image names
\usepackage{url}                % For displaying URLs
\usepackage{lipsum}             % For placeholder text

\usepackage{todonotes} % used for editing comments
\usepackage{algorithm}
\usepackage[noend]{algorithmic} % pseudocode
\usepackage{wrapfig} % enable text wrapping around figures
\usepackage{booktabs} % used for table formatting
\usepackage{tabularx} % used for table formatting
\usepackage{makecell}  % Allows line breaks in headers
\usepackage{array}      % Required for \arraybackslash
\usepackage{hyphenat}   % Allows hyphenation within words
\usepackage{subcaption}    % for sub-tables with sub-captions
\usepackage{xspace} %to add spaces after command inserted text

\reversemarginpar               % needed to make todos show up on the left.
\DeclareMathOperator*{\argmin}{arg\,min}

\newcommand{\algocomment}[1]{\hfill $\triangleright$ #1}

\newcommand{\methodName}{SEE\xspace}
\newcommand{\methodFullName}{Stable Error-seeking Exploration\xspace}

\title{Learning to Explore in Diverse Reward Settings via Temporal-Difference-Error Maximization}

\setrunningtitle{Learning to Explore in Diverse Reward Settings}

\author{Sebastian Griesbach\textsuperscript{1}, Carlo D'Eramo\textsuperscript{1,2,3}}

\emails{sebastian.griesbach@uni-wuerzburg.de, \ carlo.deramo@uni-wuerzburg.de}

\affiliations{
$^{1}$\textbf{Center for Artificial Intelligence and Data Science, University of Würzburg, Germany}\\
$^{2}$\textbf{Department of Computer Science, Technical University of Darmstadt, Germany}\\
$^{3}$\textbf{Hessian Center for Artificial Intelligence (Hessian.ai), Germany}\\
}

\contribution{
    We propose methodological approaches to tackle three identified causes of instability as mentioned above: (1.) combining the exploration and exploitation policies into a single behavior policy to bridge the far-off-policy gap; (2.) using a maximum reward update, which is agnostic towards the length of an episode; (3.) conditioning the exploration value function on current estimates of the exploitation value function to inform it about the cause of change in the TD-error target.
}
{
    Prior works investigated framing exploration as a separate optimization problem \citep{whitney_decoupled_2021, schafer_decoupled_2022} and maximizing the TD-error \citep{simmons-edler_reward_2020}. The combination of both is notoriously challenging to stabilize without altering the original optimization objective.
}
\contribution{
     We incorporate the proposed solutions resulting in \methodName, and combine it with well-established off-policy reinforcement algorithms, namely SAC \citep{tuomas_haarnoja_soft_2018} and TD3 \citep{scott_fujimoto_addressing_2018}. Our results show that the addition of \methodName maintains performance in dense reward settings and improves robustness in sparse and exploration-adverse settings, without additional hyperparameter tuning.
}
{
    We empirically compare the performances of the base algorithms and their \methodName extensions across a set of environments with variants for dense, sparse, and exploration-adverse rewards. 
}

 % Your keywords

\summary{
    Numerous heuristics and advanced approaches have been proposed for exploration in different settings for deep reinforcement learning. Noise-based exploration generally fares well with dense-shaped rewards and bonus-based exploration with sparse rewards. However, these methods usually require additional tuning to deal with undesirable reward settings by adjusting hyperparameters and noise distributions. Rewards that actively discourage exploration can pose a major challenge. This is the case if the reward function contains an action cost and no other dense signal to follow.
    We propose a novel exploration method, \methodFullName~(\methodName), that is robust across dense, sparse, and exploration-adverse reward settings. To this endeavor, we revisit the idea of maximizing the TD-error as a separate objective. Our method introduces three design choices to mitigate instability caused by: (i.) far off-policy learning, when a behavior policy is too out of distribution w.r.t. the target policy; (ii.) the conflict of interest of terminating an episode while the exploration objective follows an always positive reward signal; (iii.) the non-stationary nature of the TD-error as a target. \methodName can be combined with off-policy algorithms without modifying the optimization pipeline of the original objective. In our experimental analysis, we show that a Soft-Actor Critic agent with the addition of \methodName performs robustly across three diverse reward settings in a variety of tasks without hyperparameter adjustments.
}

\begin{document}

%\makeCover  % Create the cover page
\maketitle  % Make the title section

\begin{abstract}
    Numerous heuristics and advanced approaches have been proposed for exploration in different settings for deep reinforcement learning. Noise-based exploration generally fares well with dense-shaped rewards and bonus-based exploration with sparse rewards. However, these methods usually require additional tuning to deal with undesirable reward settings by adjusting hyperparameters and noise distributions. Rewards that actively discourage exploration, i.e., with an action cost and no other dense signal to follow, can pose a major challenge.
    We propose a novel exploration method, \methodFullName~(\methodName)\footnote{The code to reproduce all experiments is available at: \url{https://github.com/Sebastian-Griesbach/SEE}}, that is robust across dense, sparse, and exploration-adverse reward settings. To this endeavor, we revisit the idea of maximizing the TD-error as a separate objective. Our method introduces three design choices to mitigate instability caused by far-off-policy learning, the conflict of interest of maximizing the cumulative TD-error in an episodic setting, and the non-stationary nature of TD-errors. \methodName can be combined with off-policy algorithms without modifying the optimization pipeline of the original objective. In our experimental analysis, we show that a Soft-Actor Critic agent with the addition of \methodName performs robustly across three diverse reward settings in a variety of tasks without hyperparameter adjustments.
\end{abstract}

%%%%%%%%%%%%%%%%%%%%%%%%%%%%%%%%%%%%%%%%%%%%%%%%%%%%%%%%%%%%%%%%
%% Section: Introduction
%%%%%%%%%%%%%%%%%%%%%%%%%%%%%%%%%%%%%%%%%%%%%%%%%%%%%%%%%%%%%%%%

\section{Introduction}
\label{sec:introduction}
    Exploration is widely recognized as a crucial and distinctive aspect of reinforcement learning (RL). Being such a fundamental problem, a plethora of theoretical studies and empirical works have been produced over the years. Practical solutions to tackle exploration are numerous, with a prevalence of methods based on the injection of stochastic noise into policies either directly \citep{mnih_playing_2013, lillicrap_continuous_2016, scott_fujimoto_addressing_2018, fortunato_noisy_2018} or via entropy maximization \citep{tuomas_haarnoja_soft_2018, schulman_proximal_2017}. Noise-based exploration has proven to be highly effective in settings where the reward function is informative and well-shaped. For less favorable reward settings, e.g., sparse rewards, typically more targeted methods are employed. For example, bonus-based methods, also referred to as intrinsic motivation, are a class of reward-shaping mechanisms that aim to substitute sparse rewards such that the reward signal is transformed back to a shaped reward scheme again \citep{bellemare_unifying_2016, burda_exploration_2019, yiming_wang_efficient_2023}. Such methods often substitute the reward but still rely entirely on random noise for the actual action selection for exploration. The reshaped rewards can potentially dilute the original reward function of the MDP, changing the optimal policy or causing over-exploration of irrelevant state regions, which may harm performance in other reward settings \citep{taiga_bonus-based_2020}. If in a sparse reward setting one tries to enforce efficient behavior, it is reasonable to apply action costs. The resulting reward may actively discourage exploration as every action that does not immediately reach the goal is punished, adding complication to this setting. Adapting the exploration of a given RL algorithm from one setting to another often demands a costly tuning procedure, which might be infeasible in complex problems. Therefore, making RL algorithms more robustly applicable is an ongoing effort of the research community.
    
    To this endeavor, we propose \methodFullName (\methodName), a novel approach for exploration that robustly handles diverse reward settings. SEE is a directed action selection mechanism that can be combined with any off-policy RL algorithm and does not introduce relevant additional hyperparameters. Our approach separates exploration from exploitation by optimizing for two disentangled objectives and training two individual policies. On the one hand, the \textit{exploitation} objective remains the unchanged RL objective of maximizing the (discounted) cumulative reward, and, thus, it can be combined with any existing RL optimization pipeline, including reward shaping mechanisms. On the other hand, our \textit{exploration} objective maximizes the absolute temporal-difference-error (TD-error) encountered during training rollouts. The intuition behind this choice is that a high TD-error, especially in deterministic environments, indicates the presence of potentially relevant information not yet learned. We point out that decoupling exploration and exploitation into two separate objectives is not a new idea, as shown in prior works \citep{whitney_decoupled_2021, simmons-edler_reward_2020, schafer_decoupled_2022}. However, till now, it has been challenging to devise effective and stable learning procedures when maximizing the TD-error. In this paper, we identify three distinct causes of instability affecting this setting, namely (i.) far off-policy learning, when a behavior policy is too out of distribution w.r.t. the target policy; (ii.) the conflict of interest of terminating an episode while the exploration objective follows an always positive reward signal; and (iii.) the non-stationary nature of the TD-error as a target. The \textbf{contribution} of this work is the formulation of a methodological solution to tackle each of these issues, i.e., (1.) combining the exploration and exploitation policies into a single behavior policy that bridges both distributions; (2.) using a maximum reward update, which is agnostic towards the length of an episode; (3.) conditioning the exploration value function on current estimates of the exploitation value function to inform it about the cause of change in the TD-error target. We integrate these solutions into one algorithmic approach, resulting in \methodName.
    
    We empirically analyze SEE by combining it with Soft Actor-Critic (SAC) \citep{tuomas_haarnoja_soft_2018, haarnoja_soft_2019} and TD3 \citep{scott_fujimoto_addressing_2018}. We show that incorporating \methodName enhances robustness across diverse reward settings, without the need for hyperparameter tuning. Especially SAC+\methodName shows strong performance across all tested environments. Furthermore, we conduct ablation studies to analytically evince the positive effect of our proposed design choices.

%%%%%%%%%%%%%%%%%%%%%%%%%%%%%%%%%%%%%%%%%%%%%%%%%%%%%%%%%%%%%%%%
%% Section: Preliminaries
%%%%%%%%%%%%%%%%%%%%%%%%%%%%%%%%%%%%%%%%%%%%%%%%%%%%%%%%%%%%%%%%
\section{Preliminaries}
\label{sec:preliminaries}
    We consider a reinforcement learning (RL) setting \citep{sutton_reinforcement_2020} in which the environment is modeled as a Markov decision process (MDP), defined by the tuple \(\mathcal{M} = (\mathcal{S}, \mathcal{A}, \mathcal{P}, \mathcal{R}, \gamma)\), where \(\mathcal{S}\) is the state space, \(\mathcal{A}\) the action space, \(\mathcal{P}(s' | s, a)\) the state transition function, \(\mathcal{R}(s,a,s')\) the reward function, and \(\gamma \in [0,1)\) the discount factor. A stochastic policy \(\pi_\theta : \mathcal{S} \to \mathcal{P}(\mathcal{A})\) defines \(\pi_\theta(a | s)\), the probability of taking action \(a\) in state \(s\). The objective is to maximize the expected return \(J(\pi_\theta) = \mathbb{E}_{\pi_\theta} [\sum_{t=0}^\infty \gamma^t r_t]\), where the expectation is over trajectories induced by \(\pi_\theta\).  
    In the actor-critic framework, the actor represents the policy, while the critic estimates either the state-value function \(V^\pi(s) = \mathbb{E}_{\pi} [\sum_{t=0}^\infty \gamma^t r_t | s_0 = s]\) or the action-value function \(Q^\pi(s,a) = \mathbb{E}_{\pi} [r(s,a) + \gamma Q^\pi(s',a') | s,a]\). The critic is typically updated by minimizing the temporal-difference (TD) error \(\delta = r + \gamma Q^\pi(s', a') - Q^\pi(s,a)\), while the actor is updated via the policy gradient \(\nabla_\theta J(\pi_\theta) \approx \mathbb{E}_{\pi_\theta} [\nabla_\theta \log \pi_\theta(a | s) Q^\pi(s,a)]\). This framework leads to deep RL algorithms such as TD3 \citep{scott_fujimoto_addressing_2018} and SAC \citep{tuomas_haarnoja_soft_2018}, which we are using as the basis of our work.

%%%%%%%%%%%%%%%%%%%%%%%%%%%%%%%%%%%%%%%%%%%%%%%%%%%%%%%%%%%%%%%%
%% Section: Method
%%%%%%%%%%%%%%%%%%%%%%%%%%%%%%%%%%%%%%%%%%%%%%%%%%%%%%%%%%%%%%%%
\section{\methodFullName}
\label{sec:method}
\label{subsec:exploration_mdp}
    We propose an off-policy exploration method that performs robustly in different reward scenarios without hyperparameter adjustments. To this end, we use separate objectives for \textit{exploitation} and \textit{exploration}. On the one hand, the \textit{exploitation} objective remains the regular cumulative discounted reward.
    On the other hand, the \textit{exploration} objective aims to maximize the absolute value of the encountered TD-error of the \textit{exploitation} objective throughout training rollouts. This choice entails using two separate policies, one for each objective.
    It is worth noting that prior works have already established this separation of objectives \citep{whitney_decoupled_2021, simmons-edler_reward_2020, schafer_decoupled_2022}.
    However, despite its simple formulation, the exploration objective is challenging to optimize due to what we identify to be three main causes of instability in the learning process. In the following, we describe in detail these causes of instability and propose targeted solutions for each of them, eventually devising an algorithmic solution that mitigates them and enables stable and effective learning.

\subsection{Mixing of policies}
\label{subsec:mixing}
\begin{figure}[ht]
        \centering
        \begin{subfigure}[t]{0.49\textwidth}
            \centering
            \includegraphics[width=\textwidth]{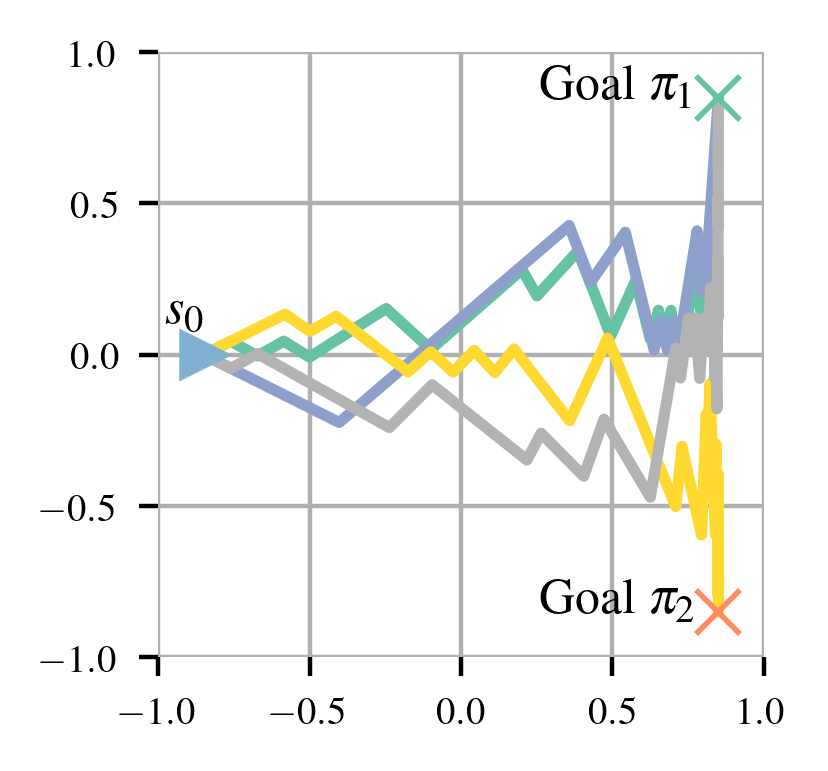}
            \caption{Four sampled trajectories of a mixed policy.}
            \label{fig:mixed_policy_trajectories}
        \end{subfigure}
        \hfill
        \begin{subfigure}[t]{0.49\textwidth}
            \centering
            \includegraphics[width=\textwidth]{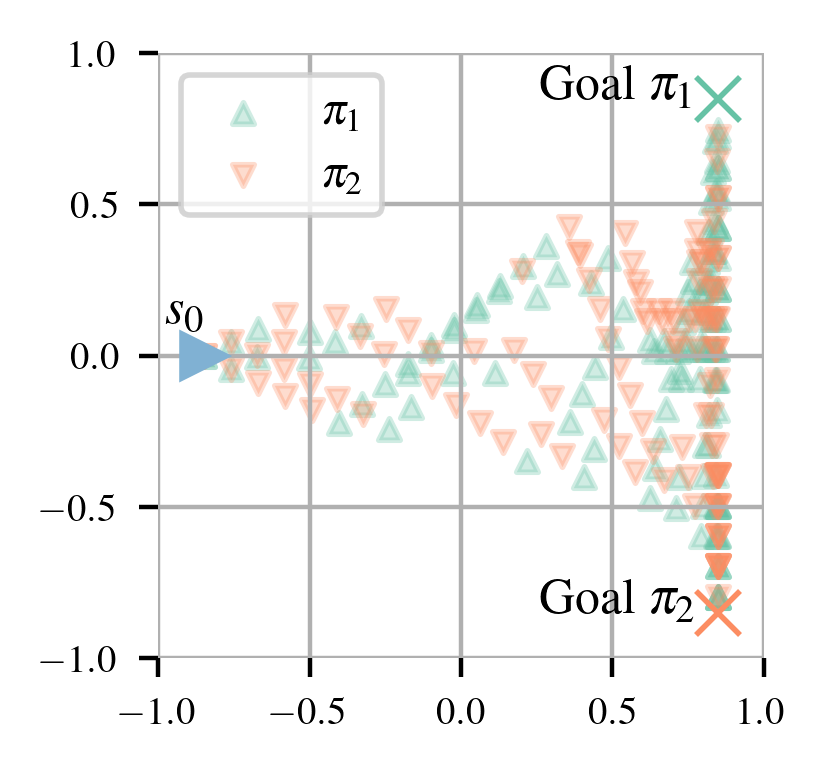}
            \caption{Distribution of states where an action of the two underlying policies is selected during the rollout.}
            \label{fig:mixed_policy_positions}
        \end{subfigure}
        \caption{Example of a mixed behavior policy of two deterministic policies $\pi_1, \pi_2$ going towards their respective goals. An action moves a maximum distance of $0.05$ on a plane of the size of $2 \times 2$. The respective action-value functions assume a reward of $1$ and termination at their target position, $0$ elsewhere. We use a discount factor of $0.9$.}
        \label{fig:mixed_policy}
    \end{figure}

    In common off-policy deep RL approaches, the behavior policy used during rollouts is chosen to be similar to the learned target policy. For example, it can be a noisy and past variant of the target policy, e.g., when using $\varepsilon$-greedy exploration with a replay buffer. Ideally, it would suffice to use an arbitrary exploration policy during training rollouts to gather relevant information about the environment. In practice, it has been shown that off-policy deep RL techniques perform poorly if the behavior policy is too out-of-distribution w.r.t. the target policy \citep{scott_fujimoto_off-policy_2019}. Importantly, by pursuing different objectives, our exploration and exploitation policies can be arbitrarily far apart, thus creating instability. To tackle this issue, we propose to combine both exploration and exploitation policies into a single \textit{behavior} policy $\mu$ which is used during training rollouts.
    Assume two arbitrary policies $\pi_1, \pi_2$ and their respective action-value functions $Q_1(s,a), Q_2(s,a)$. Let the two actions given by the policies in state $s$ be denoted as
    \begin{equation}
        a_{1} \sim \pi_{1}\left(\cdot|s\right) \hspace{2cm}
        a_{2} \sim \pi_{2}\left(\cdot|s\right).
    \end{equation}
    The behavior policy $\mu$ selects one of the candidate actions under a Boltzmann distribution
    \begin{align}
        \forall a \in \{a_1, a_2\}: \mu(a|s) &\propto \exp\left( A\left(s,a\right) \right),
    \end{align}
    where $A$ is the relative advantage of both actions w.r.t. their action-value functions defined as
    \begin{equation}
        A(s,a_1) = Q_1(s,a_1) - Q_1(s,a_2)\hspace{2cm}
        A(s,a_2) = Q_2(s,a_2) - Q_2(s,a_1).
    \end{equation}
    Using a Boltzmann distribution enables the mixing of both policies in regions where their expected values are close while focusing on one particular policy when a significantly higher value is expected. Intuitively, it explores the use of both policies and gravitates towards "goals" of any of the two. We found that a temperature of $\tau = 1$ works reliably and therefore do not consider it as a relevant hyperparameter.\\
    In general, the scale of $Q_1$ and $Q_2$ needs to be considered to achieve a balanced mixing. However, in this case, the two relative advantages are estimates of 1.) the regular advantage following a specific action which is defined as the temporal-difference (not the error) and 2.) the temporal-difference-error of the same transition. As both values are based on the same reward function, they have a similar magnitude and thus no balancing is required.
    Figure \ref{fig:mixed_policy} shows an example of a mixed policy. The data generated by the behavior policy $\mu$ creates a bridge between the two policy distributions.

\subsection{Maximum reward formulation}
\label{subsec:max_formulation}
    Typical RL approaches optimize the cumulated discounted future reward. Differently, our exploration objective optimizes the absolute value of the TD-error. This introduces a conflict of interest between the exploitation and exploration policies, especially in problems where the agent is tasked to reach a terminal state as soon as possible. The TD-error of the transition to a terminal state is likely large if the agent has not yet learned to reliably seek it. However, in the remaining state-space there might be many small TD-errors. When optimizing for the accumulated absolute TD-error, it might be suboptimal to reach the terminal state as no further rewards can be gathered thereafter. This is the same problem described by \cite{burda_exploration_2019}, who tackle it by making the exploration reward non-episodic, such that the accumulation considers the future beyond an episode termination. However, this formulation discounts the transition of interest based on its distance to the initial state, thus neglecting potentially relevant transitions that are far away from the initial state.\\
    On the contrary, we propose to use a formulation that is agnostic to the length of the episode. A first viable option would be to use an average reward formulation, as it circumvents the discounting issue altogether. Although we think this would be the most natural choice, it is not clear how to estimate the average reward of the current policy purely from off-policy data. Therefore, we opt for the maximum reward formulation where the single highest discounted reward along a trajectory is maximized \citep{gottipati_maximum_2023,veviurko_max_2024}. This simple to realize and agnostic towards the length of the episode.
    Thus, the Bellman update of the exploration objective is replaced with the maximum reward update:
    \begin{align}
        \label{eq:max_update}
        Q^{k+1}(s,a) = \max\left(r,\gamma\max_{a'}Q^k(s',a')\right),
    \end{align}
    where $r$ is the reward received during the transition from state $s$ with action $a$ to state $s'$, $\gamma$ is the discount factor, $a'$ is the next action of the underlying policy, $Q^k$ is the state-action value function and  $Q^{k+1}$ its next iteration after the update.
    Intuitively, this means that the exploration agent is seeking the single largest misconception in the value function approximation along a trajectory.

\subsection{Conditioning of value function}
\label{subsec:conditioning}
\begin{figure}[t]
    \centering
        \includegraphics[width=.75\textwidth]{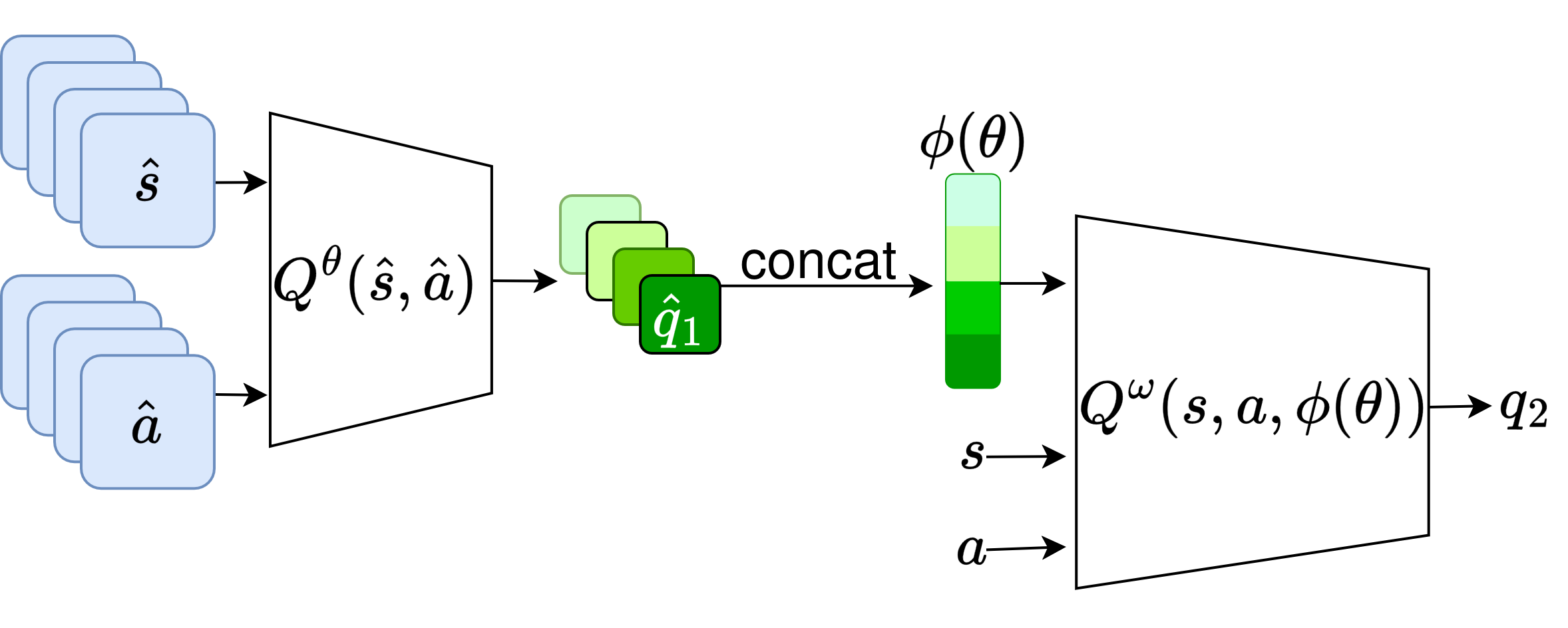}
        \caption{Fingerprinting $\phi$ is used to condition a value function $Q^\omega$ on another value function $Q^\theta$.}
        \label{fig:conditioning}
    \end{figure}

    During training, the exploitation action-value approximation changes, causing in turn the TD-error to change and thus the reward function of the exploration objective. Non-stationary rewards might destabilize training as the agent has no information on why, how, and when the reward function changes, and it is limited to adapting to the distributional shift as it appears. We propose to inform the exploration objective about the cause of changes. To achieve this, we add an embedding of the \textit{exploitation} value function parameters to the arguments of the \textit{exploration} value function.\\
    To obtain such embedding, we resort to fingerprinting, a method originally developed by \cite{harb_policy_2020}, which has been later shown to be effective in deep RL settings \citep{faccio_parameter-based_2021, faccio_general_2022}. Here, we adopt fingerprinting as follows.
    Let there be two action-value functions $Q^\theta$ and $Q^\omega$ where $\theta$ and $\omega$ are the parameters of the respective value functions. The goal is to condition $Q^\omega$ such that it takes $Q^\theta$ into account $Q^\omega(s,a|Q^\theta)$. A number of probe states $\hat{s} \in \mathcal{S}$ and probe action $\hat{a} \in \mathcal{A}$ are randomly initialized. The embedding is simply the concatenated result of passing the probe states and actions through $Q^\theta$:
    \begin{align}
        \phi(\theta) &= (Q^\theta(\hat{s}_1,\hat{a}_1), Q^\theta(\hat{s}_2,\hat{a}_2), \cdots, Q^\theta(\hat{s}_n,\hat{a}_n)),
    \end{align}
    where $\phi(\theta)$ is a vector that embeds $Q^\theta$.
    The embedding is then given as an additional input to $Q^\omega$.
    Figure \ref{fig:conditioning} shows a conceptual visualization of fingerprinting.
    As the embedding process is fully differentiable, the probe states $\hat{s}$ and actions $\hat{a}$ can be optimized based on the objective along with all other parameters. For ease of notation, the probe states $\hat{s}$ and actions $\hat{a}$ are from here on considered to be part of the conditioned value function parameters $\hat{s},\hat{a}\in \omega$. Thus, $\theta$ becomes directly an additional input $Q^\omega(s,a,\theta)$. For our method, we use fingerprinting to condition the exploration value function on the exploitation value function to inform it about the cause of change in the TD-errors. The number of used probe states and actions is an additional hyperparameter. However, \cite{faccio_general_2022} has shown that as little as 10 probe-states are sufficient to inform a critic about an actor network to solve MuJoCo environments. We do not tune for this hyperparameter but simply pick a slightly higher number of $16$ for all tasks.

\subsection{Exploration objective}
\label{subsec:exploration_objective}
    Following the three design choices described above, we can formulate our \textit{exploration} objective, which solves the MDP of the \textit{exploitation} objective with two modifications:
    \begin{itemize}
    \item The parameters of the exploitation network $\theta \in \Theta$ are included in the state space (conditioning):
    \begin{align}
        \mathcal{S}_\Delta = \mathcal{S} \cup \Theta.
    \end{align}
    \item The exploration reward function $\mathcal{R}_{\Delta}(s,a,s', \theta)$ is defined as the absolute TD-error of the exploitation action-value estimation:
    \begin{align}
    \label{eq:exploration_reward}
        \mathcal{R}_{\Delta}(s,a,s', \theta) = \left| \mathcal{R}(s,a,s') + \gamma \max_{a'} \hat{Q}^{\theta}(s',a') - \hat{Q}^\theta(s,a) \right|,
    \end{align}
    where $\mathcal{R}(s,a,s')$ is the reward function of the exploitation MDP and $\hat{Q}^\theta$ is the approximation of the exploitation state-action value function with parameters $\theta$.
    \end{itemize}
    We denote the state-action value function approximation of the \textit{exploration} objective with parameters $\omega$ as $\hat{\Delta}^\omega(s,a,\theta)$. The fingerprinting probe inputs are contained in $\omega$.
    Using the maximum update formulation and conditioning, we obtain the \textit{exploration} objective
    \begin{align}
    \label{eq:exploration_loss}
        \mathcal{J}_\omega = \mathbb{E}_{s,a,s' \sim \mathcal{D}}\left[\left(\max\left( \mathcal{R}_\Delta(s,a,s',\theta), \gamma\max_{a'}\hat{\Delta}^{\omega'}(s',a', \theta)\right) - \hat{\Delta}^\omega(s,a, \theta)\right)^2\right],
    \end{align}
    where $\mathcal{D}$ is the transition replay buffer.
    During training rollouts, a mixed behavior policy $\mu$ samples actions to collect data, as described in Section \ref{subsec:mixing}.\\
    The \textit{exploitation} objective remains the unchanged objective of the underlying base RL algorithm. It is only affected by the gathered data from the replay buffer during updates. Algorithm \ref{alg:ac+see} shows how \methodName can be combined with an arbitrary off-policy actor-critic RL method. For the sake of clarity, some details like target networks are omitted. We refer to Appendix \ref{sec:implementation} for implementation details.
    \begin{algorithm}[t]
       \caption{Generalized actor-critic + \methodName}
       \label{alg:ac+see}
    \begin{algorithmic}
       \STATE {\bfseries Input:} Regular inputs for actor-critic algorithm: $\hat{Q}^{\theta_1}: \mathcal{S} \times \mathcal{A} \rightarrow \mathbb{R}, \hat{\pi}_Q^{\theta_2}: \mathcal{S} \times \mathcal{A} \rightarrow \mathbb{R}^d$, $\theta_1 \in \Theta$;\\
       Exploration critic and actor $\hat{\Delta}^{\omega_1}: \mathcal{S} \times \mathcal{A} \times \Theta \rightarrow \mathbb{R}, \hat{\pi}_\Delta^{\omega_2}: \mathcal{S} \times \mathcal{A} \rightarrow \mathbb{R}^d$;\\
       Empty replay buffer $\mathcal{D}$;
        \FOR{each iteration}
            \FOR{each environment interaction}
                \STATE $a_Q \sim \hat{\pi}_Q^{\theta_2}(\cdot|s)$, $a_\Delta \sim \hat{\pi}_\Delta^{\omega_2}(\cdot|s)$\algocomment{sample candidate actions}
                \STATE $A(s,a_Q) \leftarrow \hat{Q}^{\theta_1}\left(s,a_Q\right) - \hat{Q}^{\theta_1}\left(s,a_\Delta\right)$ \algocomment{calculate relative advantages}
                \STATE $A(s,a_\Delta) \leftarrow \hat{\Delta}^{\omega_1}\left(s,a_\Delta, \theta_1\right) - \hat{\Delta}^{\omega_1}\left(s,a_Q, \theta_1\right)$
                \STATE $a \sim  \mu(a|s) \propto \exp(A(s,a)), \forall a \in \{a_Q,a_\Delta\}$\algocomment{sample action from behavior policy}
                \STATE $s' \sim p(\cdot|s,a)$, $\mathcal{D} \leftarrow \mathcal{D} \cup \{s,a,r,s'\}$ \algocomment{step and record transition}
         \ENDFOR
         \FOR{each gradient step}
             \STATE $  s, a, r, s' \sim \mathcal{D}$ \algocomment{sample mini-batch}
            \STATE $\theta_1 \leftarrow \argmin_{\theta_1} \mathcal{J}_{\theta_1}$, $\theta_2 \leftarrow \argmin_{\theta_2} \mathcal{J}_{\theta_2}$ \algocomment{unmodified off-policy actor-critic update}
             \STATE $a'_Q \sim  \hat{\pi}_Q^{\theta_2}(\cdot|s')$, $a'_\Delta \sim  \hat{\pi}_\Delta^{\omega_2}(\cdot|s')$ \algocomment{sample next actions}
             \STATE $r_\Delta \leftarrow \left| r + \gamma\hat{Q}^{\theta_1}(s',a'_Q) - \hat{Q}^{\theta_1}(s,a)\right|$ \algocomment{calculate exploration rewards}
             \STATE $\omega_1 \leftarrow \argmin_{\omega_1}\left( \max\left( r_\Delta, \gamma\hat{\Delta}^{\omega_1}\left(s',a'_\Delta,\theta_1\right) \right) - \hat{\Delta}^{\omega_1}\left(s,a,\theta_1\right) \right)^2$ \algocomment{maximum update}
             \STATE $\omega_2 \leftarrow \argmin_{\omega_2}\left( -\hat{\Delta}^{\omega_1}\left(s, a_\Delta \sim \hat{\pi}^{\omega_2}_\Delta\left(\cdot|s\right), \theta_1 \right) \right)$ \algocomment{update exploration actor}
         \ENDFOR
      \ENDFOR
      \STATE {\bfseries Output:} $\hat{Q}^{\theta_1}, \hat{\pi}_Q^{\theta_2}, \hat{\Delta}^{\omega_1}, \hat{\pi}_\Delta^{\omega_2}$
    \end{algorithmic}
    \end{algorithm}

%%%%%%%%%%%%%%%%%%%%%%%%%%%%%%%%%%%%%%%%%%%%%%%%%%%%%%%%%%%%%%%%
%% Section: Results
%%%%%%%%%%%%%%%%%%%%%%%%%%%%%%%%%%%%%%%%%%%%%%%%%%%%%%%%%%%%%%%%

\section{Experiments}
\label{sec:results}
We empirically validate our approach by analyzing its effect when used in off-policy deep RL algorithms. We select Twin-Delayed Deep Deterministic Policy Gradient (TD3) \citep{scott_fujimoto_addressing_2018} and Soft Actor-Critic (SAC) \citep{tuomas_haarnoja_soft_2018, haarnoja_soft_2019} as base algorithms. All additions to the regular update of these base algorithms, e.g., two critic networks and entropy maximization for SAC, are also used for the exploration objective update. This means that for SAC, both the exploration and exploitation policies perform separate entropy maximization for their respective objectives. In TD3, a static noise is added on top of the policy for exploration purposes. This static nature of the exploration noise does not align well with the idea of dynamical exploration of \methodName. Therefore, we opted to use no noise in the TD3+\methodName implementation, making the action selection of the exploration and exploitation policy deterministic. However, the behavior policy stochastically selects one of the two candidate actions. Further implementation details are described in Appendix \ref{sec:implementation}. We opt to not compare to any other methods due to the lack of comparability. We are not aware of other works that focus on robustness across different reward settings in deep RL. The most related class of algorithms is intrinsic motivation, which usually substitutes the reward with novelty bonuses for very sparse rewards, for which we do not make any claims. In all used environment variants, it is possible to occasionally find a positive reward with uniform random actions. Furthermore, our method can be combined with such reward-shaping methods, as it is agnostic to the exploitation optimization pipeline and its reward function. SEE only affects the action selection during training rollouts; thus, it is not competing with methods that augment the reward or the optimization pipeline. To our knowledge, the approach of \cite{simmons-edler_reward_2020} is the closest to our work. However, a comparison is not sensible as they employ an additional method that affects the optimization pipeline to stabilize their approach \citep{simmons-edler_q-learning_2019}.

\subsection{Comparison to base algorithms}
\begin{figure}[t]
    \centering
    \includegraphics[width=\linewidth]{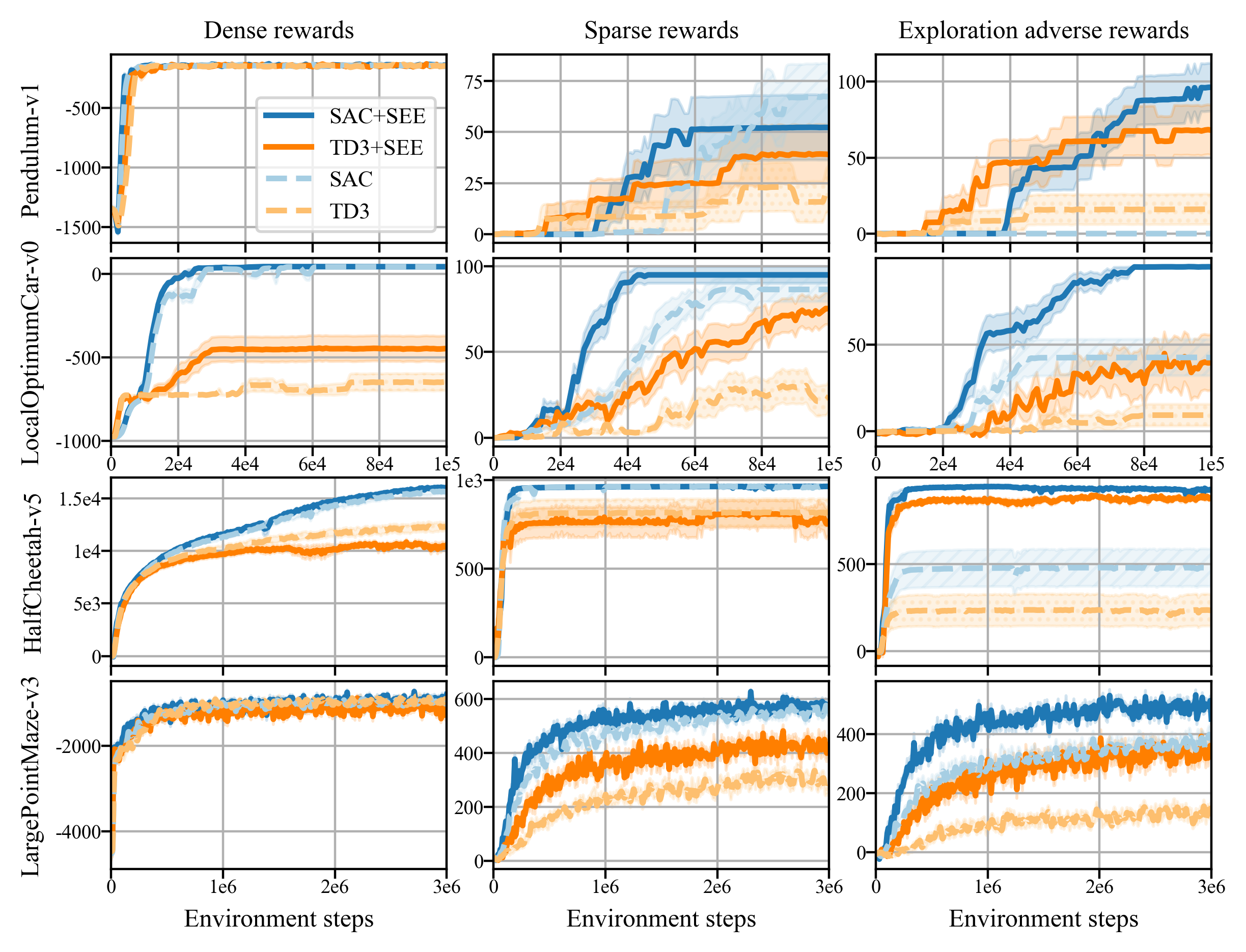}
    \caption{Comparing SAC+\methodName and TD3+\methodName to their respective base algorithms across multiple environments in different reward settings. The plots show the average evaluation return across $20$ seeds per environment variant. The shaded regions indicate the standard error.}
    \label{fig:highlighted_results}
\end{figure}

For the sake of conciseness, Figure \ref{fig:highlighted_results} highlights results of four of the total eight environments. Remaining results are shown in Figure \ref{fig:additional_results}.
We show that \methodName performs robustly in diverse reward settings by creating three reward variants per environment.
The three reward settings are dense rewards, sparse rewards, and exploration-adverse rewards. In the following, we describe what exactly these settings entail in our experimental setup.
\begin{itemize}
    \item \textbf{Dense rewards}: The reward function is well-shaped. Every step towards the desired behavior is reflected in the reward signal.
    \item \textbf{Sparse rewards}: There is a goal that must be reached. Only when the goal is reached does the reward become a positive value; otherwise, it is $0$. All goals in our environments can be occasionally reached by uniform random actions or random environment initialization.
    \item \textbf{Exploration adverse rewards}: There is a goal that must be reached. The reward function actively discourages exploration by applying an action cost to every action. A positive reward is only given upon reaching the goal. Inactivity is always a local optimum.
\end{itemize}
We select eight environments of varying difficulty from Pendulum-v1 which is quickly solved by most RL algorithms, to FetchPickAndPlace-v4 which was originally developed for Hindsight Experience Replay \citep{andrychowicz_hindsight_2017} and is quite challenging to solve without it. For each of these environments, we created three variants of the previously established reward settings. Detailed descriptions of the environments are listed in Appendix \ref{sec:environments}. For our experimental results, we run the two base algorithms and their respective \methodName extension on all $24$ environment variants. We do not change hyperparameters in between the reward settings. Notably, the hyperparameters have \textbf{not} been tuned for our extension. Instead, we use the pre-tuned hyperparameters of the base algorithms in their respective extensions. Appendix \ref{sec:hyperparameters} contains a list of all used hyperparameters.
Generally, it can be observed that in the dense reward setting, the addition of \methodName does perform comparably to their base algorithm. In the sparse reward setting, \methodName seems to have some advantage, and in the exploration adverse setting, \methodName improves performance significantly. SAC performs surprisingly well in the sparse reward setting and even has a slightly higher final performance on sparse reward Pendulum-v1. However, in the adverse reward Pendulum-v1 not a single run of SAC was capable of reaching the goal. Especially SAC+\methodName performs robustly across all environments and reward settings. TD3+\methodName generally improves the performance of TD3 with some exceptions.

\subsection{Ablations}
\label{subsec:ablations}
\begin{figure}[t]
    \centering
    \includegraphics[width=\linewidth]{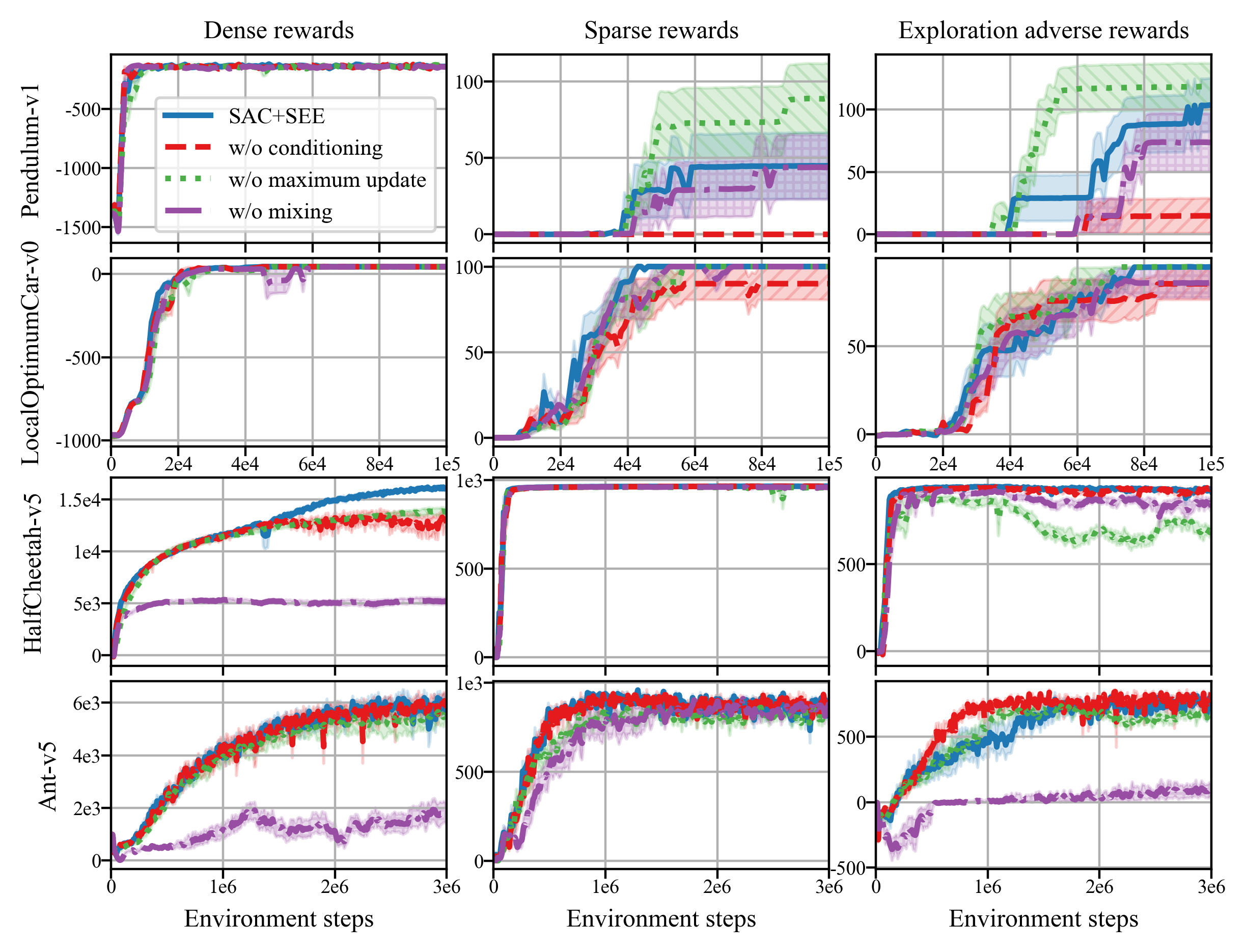}
    \caption{Comparing SAC+\methodName to ablations where one of the design choices is replaced. The graphs show the average evaluation return across $10$ seeds per environment variant. The shaded regions indicate the standard error.}
    \label{fig:sac_ablations}
\end{figure}

\begin{figure}[t]
    \centering
    \begin{subfigure}[t]{0.45\textwidth}
        \centering
        \includegraphics[width=\textwidth]{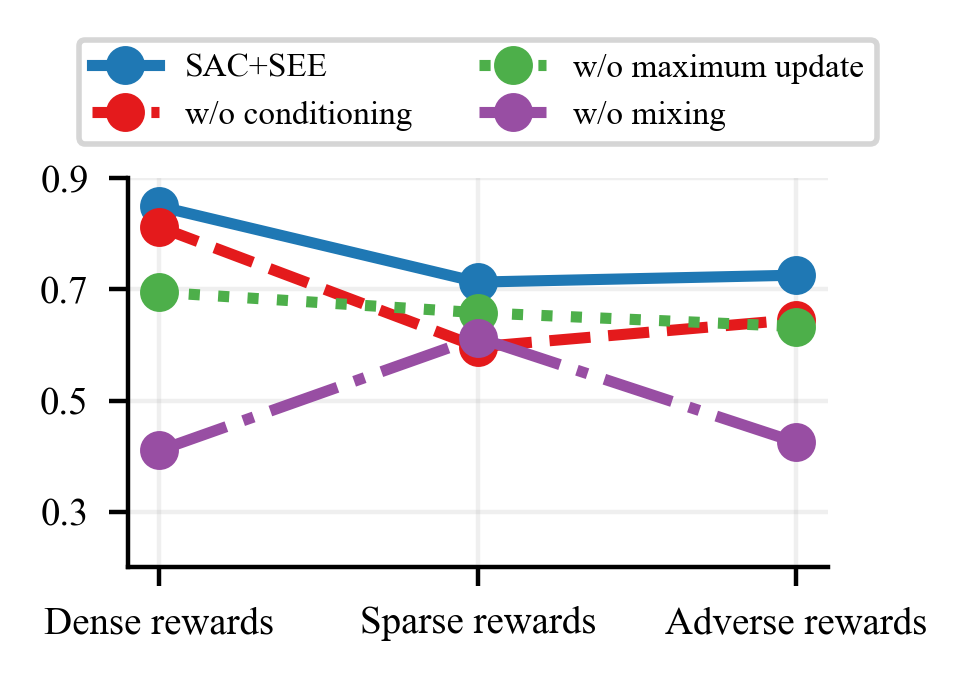}
        \caption{Accumulated average performances of SAC+\methodName ablation runs.}
        \label{fig:sac_see_ablation_profile}
    \end{subfigure}
    \hfill
    \begin{subfigure}[t]{0.45\textwidth}
        \centering
        \includegraphics[width=\textwidth]{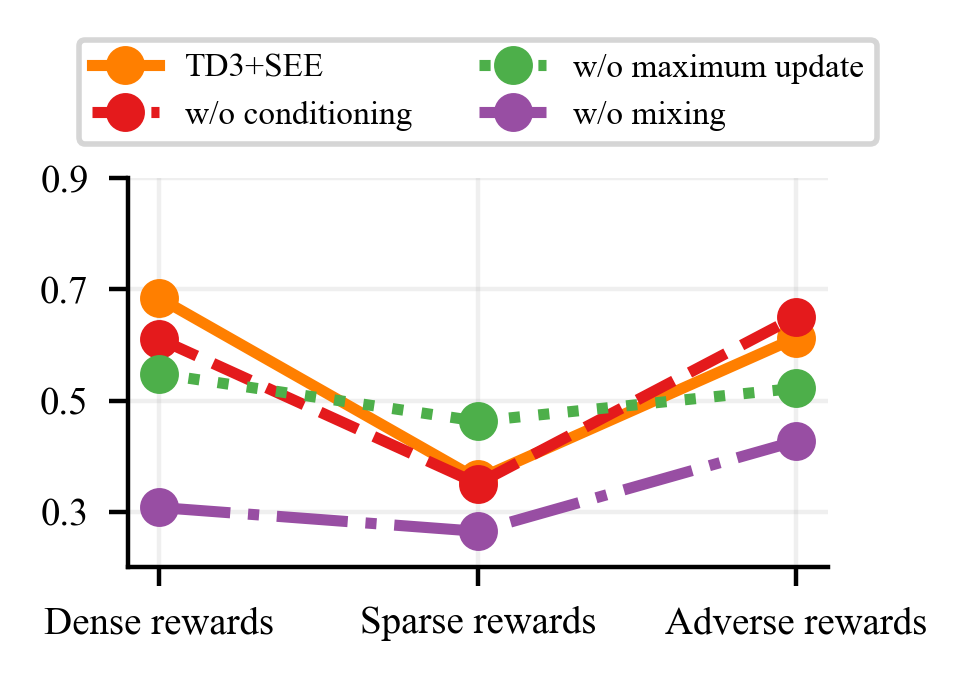}
        \caption{Accumulated average performances of TD3+\methodName ablation runs.}
        \label{fig:td3_see_ablation_profile}
    \end{subfigure}
    \caption{The graphs show the normalized aggregated average returns grouped by reward setting for all ablations. The normalization assigns the value $0$ to the single worst run in an environment variant and $1$ to the single best run.}
    \label{fig:ablation_profile}
\end{figure}

To measure the impact of our three design choices, we conducted ablation studies on a subset of environments. To avoid cluttering, we highlight in Figure \ref{fig:sac_ablations} the ablation studies of combining \methodName with SAC.
The results of the same ablation but with TD3+\methodName are presented in Figure \ref{fig:td3_ablations}.
The three conducted ablations replace one of the three design choices each.
\begin{itemize}
    \item \textbf{w/o conditioning:} The exploration value function does not receive an embedding of the exploitation value function.
    \item \textbf{w/o maximum update:} The update of the exploration objective uses a regular Bellman update instead of the maximum update.
    \item \textbf{w/o mixing:} Actions during training rollouts are selected alternately by the exploitation and exploration policy instead of using the proposed behavior policy.
\end{itemize}
We observe that some additions only have a positive effect in specific settings. The maximum update even decreases performance in the Pendulum-v1 environment. However, the combination of all three additions always performs robustly without a failure case. To evince this, Figure \ref{fig:ablation_profile} shows normalized accumulated results across all environments grouped by reward setting.
SAC+\methodName performs best on average with all active design choices.
However, this is not the case for TD3+\methodName. Here, especially the maximum update has a negative impact on performance in sparse reward settings. The most positively impactful addition is the mixing of exploration and exploitation policy.

%%%%%%%%%%%%%%%%%%%%%%%%%%%%%%%%%%%%%%%%%%%%%%%%%%%%%%%%%%%%%%%%
%% Section: Related
%%%%%%%%%%%%%%%%%%%%%%%%%%%%%%%%%%%%%%%%%%%%%%%%%%%%%%%%%%%%%%%%
\section{Related works}
\label{sec:related_works}
Exploration in RL is an active field of research. While naive undirected exploration methods like $\varepsilon$-greedy or random noise are effective when exposed to well-shaped dense rewards, they often struggle in other settings. One common idea to mitigate this effect is to substitute unfavorably shaped rewards, such that the resulting reward ends up being well-shaped again. To this end, bonus-based methods, also referred to as intrinsic reward methods, are used. While we do not think that it is sensible to compare our method to intrinsic motivation methods, we highlight some works in that field that are conceptually related to this work. Random network distillation (RND) \citep{burda_exploration_2019} popularized the idea of using a prediction error as a bonus reward. They also encountered the conflict of interests of strictly positive exploration rewards in an episodic setting. Many similar ideas followed, addressing several issues of this method. One general issue of bonus-based exploration is that the bonus objective is changing the MDP. A changed reward function also potentially changes the optimal policy. \cite{yiming_wang_efficient_2023} propose a method that substitutes sparse rewards without changing the optimal policy. For this, they employ a potential function that measures the distance of states where the distance is a bisimulation metric. Intuitively, this means that the distance of states is measured as their distance in the value space instead of the feature space. As a result, their method is more likely to uncover states with high TD-errors. Due to the bisimulation metric relying on the current policy and the respective value function, this method is limited to on-policy settings.
\cite{sukhija_maxinforl_2025} propose an intrinsic motivation method, where the bonus objective is defined as the upper bound of an information gain approximation. Their method uses a learned world model for this approximation and can be combined with any off-policy reinforcement learning method. They integrate it into SAC in the same way as the entropy maximization is and are able to show strong empirical results in sparse and dense reward environments.
The previously mentioned methods disregard the non-stationary nature of their exploration bonus. \cite{whitney_decoupled_2021} recognize this and decouples exploration from exploitation. The exploration agent here follows an optimistic pseudo-count intrinsic reward. To better deal with the non-stationarity, the exploration reward is calculated during the update and not stored in the replay buffer. Additionally, the exploration agent is updated more aggressively than the exploitation agent. To overcome the far-off-policy instability, they use a product of exploration and exploitation policy as a behavior policy, which has the downside that these policies require a significant overlap to work in practice. \cite{simmons-edler_reward_2020} is similar to our method; they defined a separate exploration objective for maximizing the TD-error incurred by the exploitation objective. To solve the far-off-policy instability, they roll out both policies individually and collect the data in separate buffers. During the update, the data is mixed with a specific ratio. They had trouble stabilizing this approach with DDPG-style parametric policies and therefore augmented the optimization pipeline with an additional method \citep{simmons-edler_q-learning_2019}.
Our approach roughly follows the idea of \citet{riedmiller_collect_2021} who propose to see exploration and exploitation as two separate phases in the training refereed to as 'Collect and Infer'. During the collection phase, the objective is to collect the optimal dataset such that during the inference phase, the best possible policy based on that amount of data can be learned. This requires the data collection to be aware of what data has already been collected. While our method still balances exploration and exploitation during training, it is informed about collected data through the conditioning.

%%%%%%%%%%%%%%%%%%%%%%%%%%%%%%%%%%%%%%%%%%%%%%%%%%%%%%%%%%%%%%%%
%% Section: Discussion
%%%%%%%%%%%%%%%%%%%%%%%%%%%%%%%%%%%%%%%%%%%%%%%%%%%%%%%%%%%%%%%%

\section{Discussion}
\label{sec:discussion}
After empirically showing the strength of our proposed method \methodName, we also want to discuss limitations of the approach. Due to the additional exploration objective, we double the amount of required function approximations. Thus, in turn with the additional updates, roughly doubling the required compute compared to the base algorithm.
Bonus-based exploration methods often suffer from the so-called noisy-TV problem. It describes a setting where a novelty-seeking agent gets stuck in stochastic transitions that constantly produce novel states, such as a noisy TV. \methodName is not directly affected by this as it does not seek novel states; however, stochastic rewards may produce a similar effect. For some transitions with a stochastic reward, the exploitation agent might learn the expected reward, but due to the stochasticity, there will be a consistent TD-error related to this transition on which the exploration policy might get stuck on.\\
We have shown that SAC+\methodName works reliably in a diverse set of environments and reward settings, notably without specifically tuning hyperparameters. Therefore, we consider SAC+\methodName an interesting candidate for challenging reward settings where precise hyperparameter tuning might not be feasible. Furthermore, we think that the individual design choices may also be applicable in other settings, such as when a behavior policy needs to reflect multiple objectives (mixing), a decision-making agent should be indifferent to the length of an episode (maximum update), or where an objective depends on a non-stationary target (conditioning).

% \subsubsection*{Broader Impact Statement}
% \label{sec:broaderImpact}
% In this optional section, RLJ/RLC encourages authors to discuss possible repercussions of their work, notably any potential negative impact that a user of this research should be aware of. 
% \todo{Talk about danger of risk seeking behavior}

%%%%%%%%%%%%%%%%%%%%%%%%%%%%%%%%%%%%%%%%%%%%%%%%%%%%%%%%%%%%%%%%
%% Appendices
%%%%%%%%%%%%%%%%%%%%%%%%%%%%%%%%%%%%%%%%%%%%%%%%%%%%%%%%%%%%%%%%
\newpage

\appendix

\section{Additional results}
\label{sec:additional_results}
\begin{figure}[t]
    \centering
    \includegraphics[width=\linewidth]{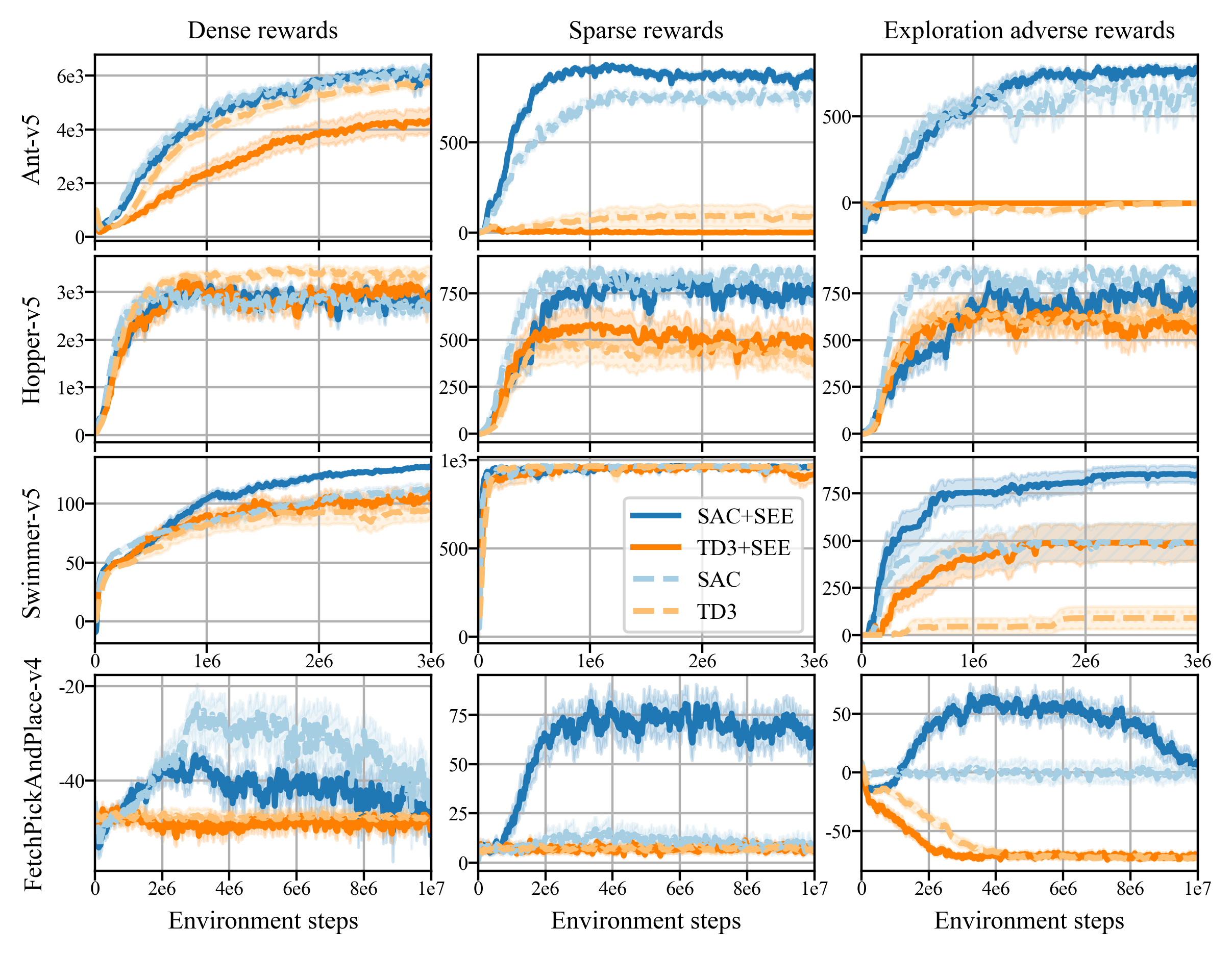}
    \caption{Comparing SAC+\methodName and TD3+\methodName to their respective base algorithms across multiple environments with different reward settings. The graphs show the average evaluation return across $20$ seeds per environment variant. The shaded regions indicate the standard error.}
    \label{fig:additional_results}
\end{figure}

\begin{figure}[t]
    \centering
    \includegraphics[width=\linewidth]{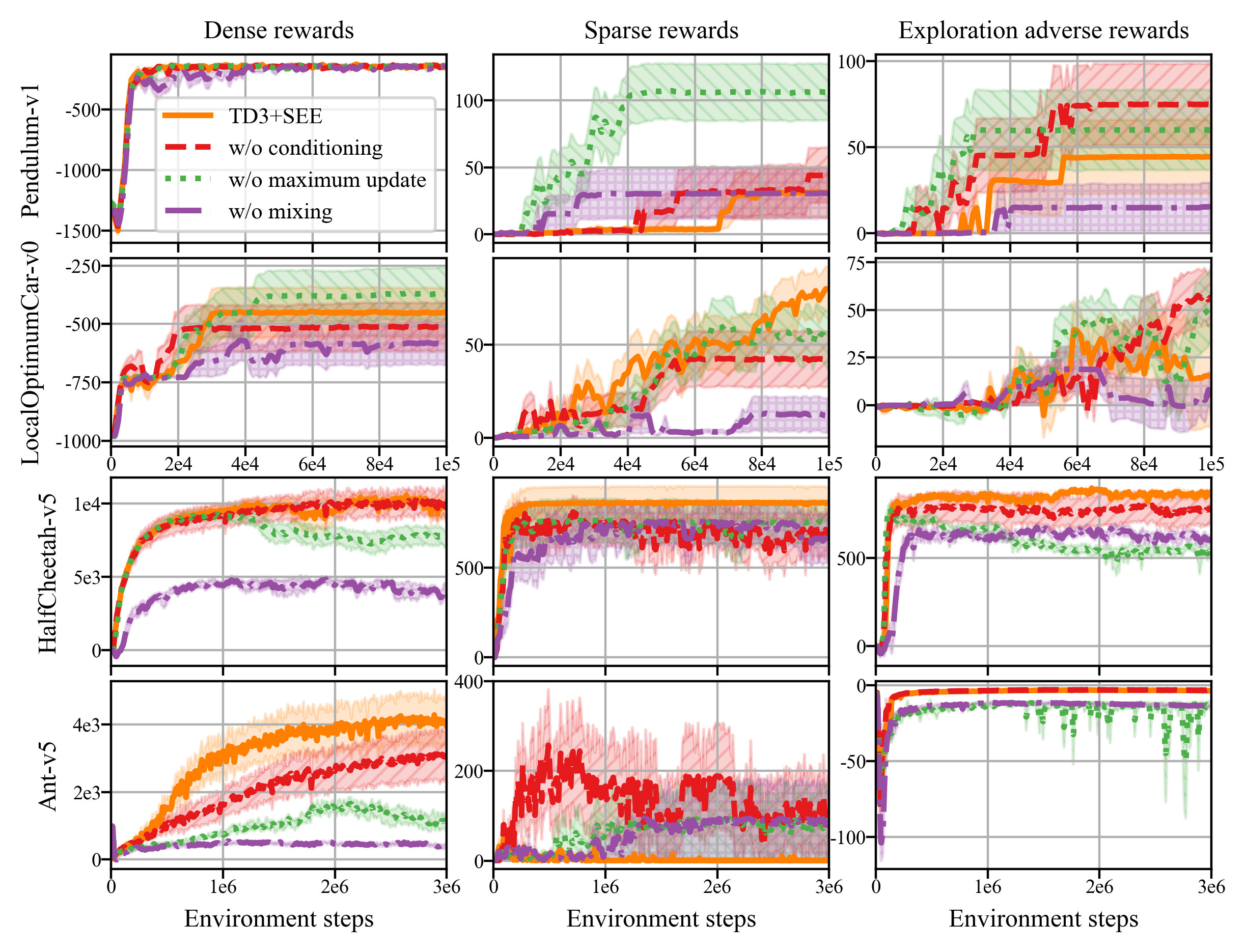}
    \caption{Comparing TD3+\methodName to ablations where one of the design choices is replaced. The graphs show the average evaluation return across $10$ seeds per environment variant. The shaded regions indicate the standard error.}
    \label{fig:td3_ablations}
\end{figure}

In this appendix we present the omitted results from the main paper for the sake of conciseness. Figure \ref{fig:additional_results} depicts the remaining environments of the comparison of \methodName to the base algorithms.
Noteworthy is that in the Hopper-v5 environment, the addition of \methodName is unable to improve performance even in difficult reward settings. In Swimmer-v5 SAC+\methodName outperforms SAC in the dense reward setting. We think this is due to the fact that usually for this environment a high discount of $\gamma = 0.9999$ is used. However, in our experiments, we use the same hyperparameters for all Mujoco environments (see Appendix \ref{sec:hyperparameters}) and \methodName seems to perform robustly in the absence of specific tuning. In the sparse FetchPickAndPlace-v4, all methods regularly encounter a positive reward as the box sometimes is initialized on top of the target position. But only SAC+\methodName was capable of picking up on that signal.\\
Figure \ref{fig:td3_ablations} shows the individual ablation studies of TD3+\methodName.
It can be seen that in combination with TD3 these additions are not as effective. As previously mentioned, for TD3+\methodName we do not add stochastic noise to our actions. This result therefore might hint towards stochastic exploration being beneficial even in the presence of a dedicated exploration objective. As a reminder, in SAC+\methodName we employ entropy maximization on both the exploitation and exploration objectives.

% \noindent \textbf{Note:} Appendices appear before the references and are viewed as part of the ``main text'' and are subject to the 8--12 page limit, are peer reviewed, and can contain content central to the claims of the paper. 

% \section*{Appendix}
% % No label, since this can't be referenced meaningfully with \ref{}.
% This format should only be used if there is a single appendix (unlike in this document).

\newpage

\subsubsection*{Acknowledgments}
\label{sec:ack}
This work was funded by the German Federal Ministry of Education and Research (BMBF) (Project: 01IS22078). 

The authors gratefully acknowledge the scientific support and HPC resources provided by the Erlangen National High Performance Computing Center (NHR@FAU)
of the Friedrich-Alexander-Universität Erlangen-Nürnberg (FAU) under the NHR project b187cb. NHR funding is provided by federal and Bavarian state authorities. NHR@FAU hardware is partially funded by the German Research Foundation (DFG) – 440719683.

The authors also gratefully acknowledge the "Julia 2" HPC provided by the Julius-Maximilians-University Würzburg. "Julia 2" was funded as DFG project as "Forschungsgroßgerät nach Art 91b GG" under INST 93/1145-1 FUGG.

%%%%%%%%%%%%%%%%%%%%%%%%%%%%%%%%%%%%%%%%%%%%%%%%%%%%%%%%%%%%%%%%
%% NOTE: THIS MARKS THE END OF THE "MAIN TEXT"
%%%%%%%%%%%%%%%%%%%%%%%%%%%%%%%%%%%%%%%%%%%%%%%%%%%%%%%%%%%%%%%%

%%%%%%%%%%%%%%%%%%%%%%%%%%%%%%%%%%%%%%%%%%%%%%%%%%%%%%%%%%%%%%%%
%% Bibliography
%%%%%%%%%%%%%%%%%%%%%%%%%%%%%%%%%%%%%%%%%%%%%%%%%%%%%%%%%%%%%%%%
% \bibliography{main}
\clearpage
\bibliography{references}
\bibliographystyle{rlj}

%%%%%%%%%%%%%%%%%%%%%%%%%%%%%%%%%%%%%%%%%%%%%%%%%%%%%%%%%%%%%%%%
% AUTHOR: If your paper has no supplementary materials, you may 
%         comment out the line below, which creates the title for
%         the supplementary materials.
%%%%%%%%%%%%%%%%%%%%%%%%%%%%%%%%%%%%%%%%%%%%%%%%%%%%%%%%%%%%%%%%
\beginSupplementaryMaterials

\section{Environments}
\label{sec:environments}
In this work we use modified versions of well know environments.
Table \ref{tab:environments} gives details on the modification that was done for each environment.
\begin{table}[ht]
    \centering
    \renewcommand{\arraystretch}{1.2} % Adjust row height for readability
    {\small
    \begin{tabularx}{\textwidth}{|>{\raggedright\arraybackslash}X|>{\raggedright\arraybackslash}X|>{\raggedright\arraybackslash}X|>{\raggedright\arraybackslash}X|>{\raggedright\arraybackslash}X|}
        \hline
        \textbf{\shortstack{Name}} & \textbf{\shortstack{General}} & \textbf{\shortstack{Dense}} & \textbf{\shortstack{Sparse}} & \textbf{\shortstack{Adverse}} \\ 
        \hline
        \href{https://gymnasium.farama.org/environments/classic_control/pendulum/}{Pendulum-v1} & unmodified & unmodified & +1 if Pendulum within upper 10 degree else 0, Pendulum always starts pointing down with 0 velocity & sparse reward + action cost of unmodified version \\ 
        \hline
        LocalOptimum-Car-v0 & \href{https://gymnasium.farama.org/environments/classic_control/mountain_car_continuous/}{MountainCar-Continuous-v0} with an additional lesser goal on the left at position –1.1 giving a reward of +10 (see Figure \ref{fig:localOptimumCar}) & Original reward with additional negative distance to the optimal goal & original reward without action cost & original reward \\ 
        \hline
        \href{https://gymnasium.farama.org/environments/mujoco/half_cheetah/}{HalfCheetah-v5} & for sparse and adverse setting observation includes position of the agent & unmodified & +1 for x-position > 5 else 0 & sparse reward + action cost of unmodified version \\ 
        \hline
        \href{https://gymnasium.farama.org/environments/mujoco/ant/}{Ant-v5} & for sparse and adverse setting observation includes position of the agent & unmodified & +1 for x-position > 4 else 0 & sparse reward + action cost of unmodified version \\ 
        \hline
        \href{https://gymnasium.farama.org/environments/mujoco/hopper/}{Hopper-v5} & for sparse and adverse setting observation includes position of the agent & unmodified & +1 for x-position > 1 else 0 & sparse reward + action cost of unmodified version \\ 
        \hline
        \href{https://gymnasium.farama.org/environments/mujoco/swimmer/}{Swimmer-v5} & for sparse and adverse setting observation includes position of the agent & unmodified & +1 for x-position > 1 else 0 & sparse reward + action cost of unmodified version \\
        \hline
        \href{https://robotics.farama.org/envs/fetch/pick_and_place/}{FetchPickAnd-Place-v4} & Adding goal position to observation, episode truncated at $200$ steps & using reward of FetchPickAnd-PlaceDense-v4 & $+1$ if distance to goal $<= 0.05$ else $0$  & sparse reward - $0.1\sum^{dim(\mathcal{A})}_{i=0}a_i^2$ \\
        \hline
        LargePoint-Maze-v3 & \href{https://robotics.farama.org/envs/maze/point_maze/}{PointMaze\_Large\_Diverse\_GR-v3}, adding goal position to observation & reward is negative distance to goal position & $+1$ if distance to goal $<= 0.45$ else $0$ & sparse reward - $0.1\sum^{dim(\mathcal{A})}_{i=0}a_i^2$ \\
        \hline
    \end{tabularx}
    }
    \caption{List of all used environments with descriptions of general modifications and modifications made for each of the three setting namely dense-rewards, sparse-rewards and exploration adverse-rewards.}
    \label{tab:environments}
\end{table}
All environments are original or modified versions of the environments found in \href{https://gymnasium.farama.org/}{Gymnasium} \citep{towers_gymnasium_2023} and \href{https://robotics.farama.org/#}{Gymnasium-Robotics}. Refer to \href{https://github.com/Sebastian-Griesbach/SEE}{our provided code} for more details.
\begin{figure}[ht]
    \centering
    \includegraphics[width=0.5\linewidth]{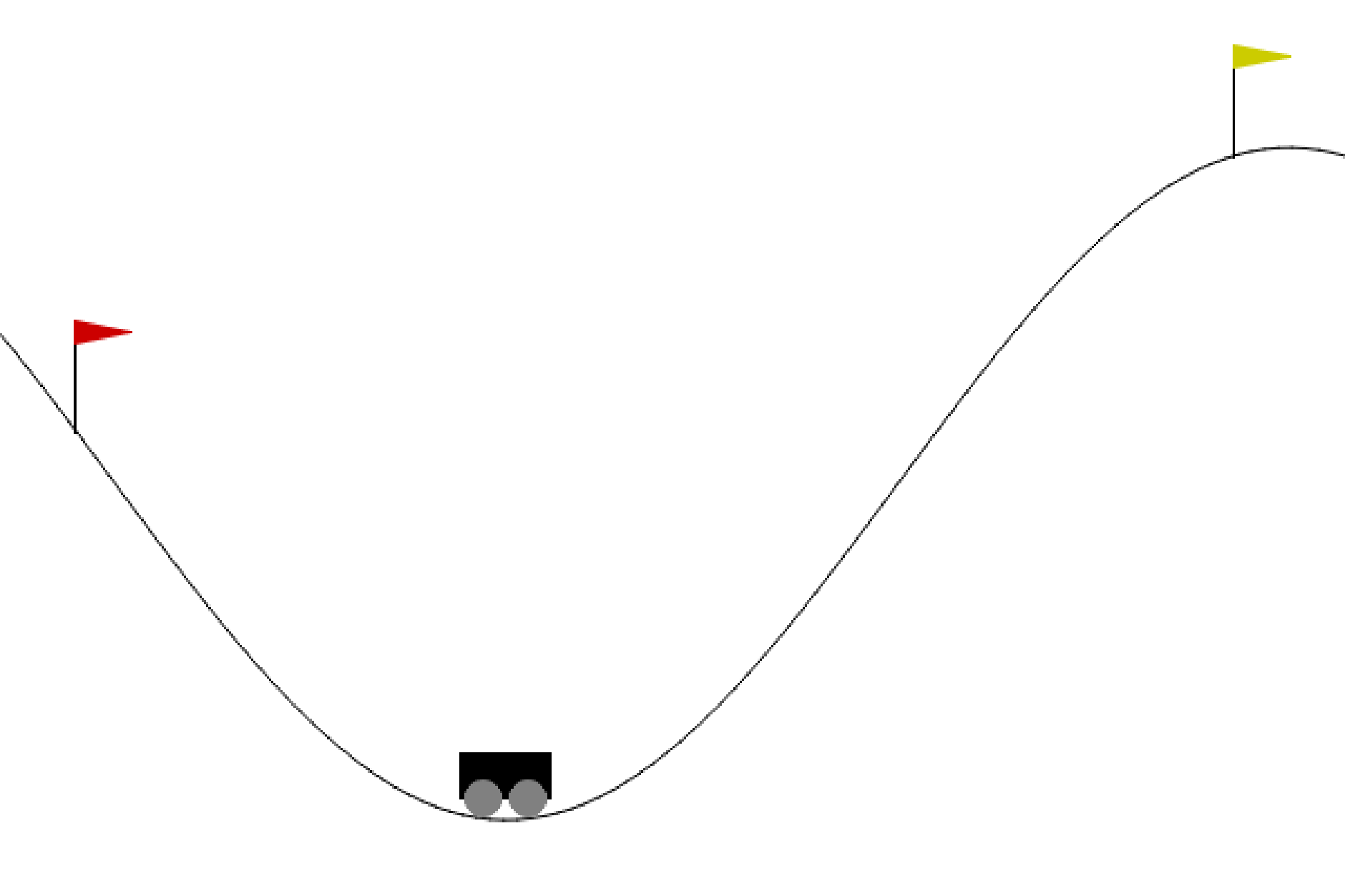}
    \caption{Depiction of LocalOptimumCar-v0 a modified version of MountainCarContinuous-v0 with an additional easier to reach goal on the left. Unlike the goal on the right the additional goal only yields a reward of $10$ when reached instead of $100$.}
    \label{fig:localOptimumCar}
\end{figure}

\section{Hyperparameters}
\label{sec:hyperparameters}
Table \ref{tab:hyperparam_overview} shows used hyper parameters. 
\begin{table}[ht]
    \centering
    \begin{tabular}{c@{\hspace{1cm}}c@{\hspace{1cm}}c@{}}
        % First row: two subtables
        \begin{subtable}[t]{0.48\textwidth}
            \centering
            \begin{tabular}{lcc}
                \toprule
                \makecell{\textbf{Hyperparameter}} & \makecell{\textbf{Classic}} & \makecell{\textbf{MuJoCo}} \\
                \midrule
                Learning rate & 0.001 & 0.0003 \\
                Batch size    & 256   & 256    \\
                Discount $\gamma$ & 0.99 & 0.99 \\
                Replay buffer size & $200000$ & $1000000$ \\
                Target networks $\tau$ & $0.005$ & $0.005$ \\
                Update freq & $1$ & $1$ \\
                Warm-up steps & $1000$ & $10000$ \\
                Hidden dims & $[400, 300]$ & $[400, 300]$ \\
                Temperature & auto & auto \\
                Initial temperature & $1.0$ & $1.0$ \\
                Target entropy & auto & auto \\
                \bottomrule
            \end{tabular}
            \caption{Hyperparameters used for SAC}
            \label{tab:hyper_sac}
        \end{subtable}
        &
        \begin{subtable}[t]{0.48\textwidth}
            \centering
            \begin{tabular}{lcc}
                \toprule
                \makecell{\textbf{Hyperparameter}} & \makecell{\textbf{Classic}} & \makecell{\textbf{MuJoCo}} \\
                \midrule
                Learning rate & 0.001 & 0.001 \\
                Batch size    & 256   & 256   \\
                Discount $\gamma$ & 0.99 & 0.99 \\
                Replay buffer size & $200000$ & $1000000$ \\
                Target networks $\tau$ & $0.005$ & $0.005$ \\
                Critic update freq & $1$ & $1$ \\
                Actor update freq & $2$ & $2$ \\
                Target update freq & $2$ & $2$ \\
                Warm-up steps & $1000$ & $10000$ \\
                Hidden dims & $[400, 300]$ & $[400, 300]$ \\
                Action noise std & $0.1$ & $0.1$ \\
                Target noise std & $0.2$ & $0.2$ \\
                Target noise clip & $0.5$ & $0.5$ \\
                \bottomrule
            \end{tabular}
            \caption{Hyperparameters used for TD3}
            \label{tab:hyper_td3}
        \end{subtable} 
        \\[12em]
        % Second row: two subtables
        \begin{subtable}[t]{0.48\textwidth}
            \centering
            \begin{tabular}{lcc}
                \toprule
                \makecell{\textbf{Hyperparameter}} & \makecell{\textbf{Classic}} & \makecell{\textbf{MuJoCo}} \\
                \midrule
                Probe states (\ref{subsec:conditioning}) & $16$ & $16$ \\
                \bottomrule
            \end{tabular}
            \caption{Hyperparameters used for SAC+\methodName that differ from SAC}
            \label{tab:hyper_td3+see}
        \end{subtable}
        &
        \begin{subtable}[t]{0.48\textwidth}
            \centering
            \begin{tabular}{lcc}
                \toprule
                \makecell{\textbf{Hyperparameter}} & \makecell{\textbf{Classic}} & \makecell{\textbf{MuJoCo}} \\
                \midrule
                Action noise std & $0.0$ & $0.0$ \\
                Probe states (\ref{subsec:conditioning}) & $16$ & $16$ \\
                \bottomrule
            \end{tabular}
            \caption{Hyperparameters used for TD3+\methodName that differ from TD3}
            \label{tab:hyper_sac+see}
        \end{subtable}
    \end{tabular}
        \caption{Overview of used hyperparameters by methods and environments. Classic includes the environment Pendulum-v1 and LocalOptimumCar-v0. MuJoCo includes all remaining environments. For the FetchPickAndPlace-v4 environment all methods used one additional hidden layer and thus had hidden dims of $[400,300,200]$.}
    \label{tab:hyperparam_overview}
\end{table}
These were taken from the \href{https://github.com/DLR-RM/rl-baselines3-zoo}{RL Baseline3 Zoo} \citep{raffin_rl_2020}. The parameter sets for the Classic environments were taken from the tuned version of Pendulum-v1 for TD3 and SAC respectively, and the MuJoCo sets were taken from HalfCheetah-v4 hyperparameters. The discount factor is always set to $0.99$. \methodName does not introduce relevant new hyperparameters but doubles them as we separately optimize for two objectives. All hyperparameters for the \textit{exploration} objective are copied from the \textit{exploitation} objective. For the FetchPickAndPlace environments, we use an additional hidden layer such that the hidden dims are $[400,300,200]$. All remaining hyperparameters are the same as the MuJoCo setting.

\section{Implementation details}
\label{sec:implementation}
For the sake of clarity, some details of \methodName are omitted from the main paper that concern the implementation.
\paragraph{Policy mixing balance} In section \ref{subsec:mixing} it is mentioned that in our case the two objectives do not require scaling to achieve a balanced mixing due to a similar magnitude. In practice we use two mixing factors $\lambda$ and $(1-\lambda)$ for the relative advantage values during mixing. For all our experiments lambda is set to $\lambda = 0.5$. Therefore, this does not change their relative magnitude. However, it does change their absolute magnitude, which could affect the sampling probabilities of the Boltzmann distribution.
\paragraph{Exploration Update} The \methodName extensions make use of all update methods used in the respective base algorithms. Including target networks, multiple critics, soft critics, minimum over multiple critics for target calculation, etc. This means that if the base algorithm uses the minimum over two target networks for its next value estimate, so does the update of our exploration objective.
\paragraph{Relative advantage calculation} To calculate the relative advantage, we take the difference of two state-action values. The exact calculation of these state-action values imitates the calculation of the state-action value for the actor update of the base algorithm. This means that in TD3+\methodName this is simply the state-action value given by the first of the two critic networks. In SAC+\methodName this is the minimum across both state-action values.\\
\paragraph{Use of online parameters for exploration reward} For calculating the exploration reward, only the online parameters of the exploitation objective are used (not of the target network), as these are also the only parameters used for the conditioning of the exploration objective.\\

\paragraph{Reproducibility}
Our full implementation to reproduce all results is available at:\\ \url{https://github.com/Sebastian-Griesbach/SEE}.

\end{document}